\documentclass[letterpaper, 10 pt, conference]{ieeeconf}  

\IEEEoverridecommandlockouts                              

\usepackage{graphics} 
\usepackage{epsfig} 
\usepackage{times} 
\usepackage{amsmath} 
\usepackage{amssymb}  
\usepackage{algorithm}
\PassOptionsToPackage{noend}{algpseudocode}
\usepackage{algpseudocode}
\usepackage[font=small]{caption}
\usepackage{subcaption}
\usepackage{units}

\usepackage{fontenc}
\usepackage{bm}
\usepackage[utf8]{inputenc}
\usepackage{mathtools}
\usepackage{color}

\usepackage{epstopdf}
\usepackage{tabularx}
\usepackage{hhline}
\usepackage{url}
\usepackage{leftidx}
\usepackage{tensor}

\usepackage[nolist,nohyperlinks]{acronym}
\usepackage{enumerate}
\usepackage{todonotes}
\usepackage{graphicx}
\usepackage{caption}
\usepackage{subcaption}
\usepackage{gensymb}
\usepackage{booktabs}
\usepackage{float}

\usepackage{amssymb}
\usepackage{amsmath}
\usepackage{multicol}

\title{\LARGE \bf
	Nonlinear Model Predictive Control for Multi-Micro Aerial Vehicle Robust Collision Avoidance
}

\author{Mina Kamel\authorrefmark{1}, Javier Alonso-Mora\authorrefmark{2}, Roland Siegwart\authorrefmark{1}, and Juan Nieto\authorrefmark{1}\\
	\authorblockA{\authorrefmark{1}Authors are with the Autonomous Systems Lab, ETH Zurich}
	 \authorrefmark{2}Author is with the Delft Center for Systems and Control, TU Delft}

\begin{document}

	\maketitle
	\thispagestyle{empty}
	\pagestyle{empty}

	\graphicspath{{Figures/}}
	\begin{abstract}
		%
		Multiple multirotor \acp{MAV} sharing the same airspace require a reliable and robust collision avoidance technique. In this paper we address the problem of multi-\ac{MAV} reactive collision avoidance. A model-based controller is employed to achieve simultaneously reference trajectory tracking and collision avoidance. Moreover, we also account for the uncertainty of  the state estimator and the other agents position and velocity uncertainties to achieve a higher degree of robustness. The proposed approach is decentralized, does not require collision-free reference trajectory and accounts for the full \ac{MAV} dynamics. We validated our approach in simulation and experimentally.
	\end{abstract}


	\section{INTRODUCTION}\label{sec:intro}
	As the miniaturization technology advances, low cost and reliable \acp{MAV} are becoming available on the market with powerful on-board computation power. Many applications can benefit from the presence of such low cost systems such as inspection and exploration~\cite{treecavity2016, nbvp2016}, surveillance for security~\cite{BorderPatroUAV1}, mapping~\cite{Oettersagen_FSR_15} or crop monitoring~\cite{CropMonitoringUAV1}. However, \acp{MAV} are limited to short flight times due to battery limitation and size constraints. Due to this limitation, creating a team of \acp{MAV} that can safely share the airspace to execute a specific mission will widen the range of applications where \acp{MAV} can be used and will be beneficial for time-critical missions such as search and rescue operations~~\cite{Doherty_VictimDetection}.

A crucial problem when multiple \acp{MAV} share the same airspace is the risk of mid-air collision, because of this, a robust method to avoid multi-\acp{MAV} collisions is necessary. Typically, this problem is solved by planning collision-free trajectory for each agent in a centralized manner. However, this binds the \acp{MAV} to the pre-planned trajectory and limits the adaptivity of the team during the mission: any change in the task will require trajectory re-planning for the whole team.

In this work we present a unified framework to achieve reference trajectory tracking and multi-agent reactive collision avoidance.  The proposed approach exploits the full \ac{MAV} dynamics and takes into account the physical platform limitations. In this way we fully exploit the \ac{MAV} capabilities and achieve agile and natural avoidance maneuvers compared to classic approaches, where planning is decoupled from trajectory tracking control. To this end, we formulate the control problem as a constrained optimization problem that we solve in a receding horizon fashion. The cost function of the optimization problem includes a potential field-like term that penalizes collisions between agents. While potential field methods do not provide any guarantee and are sensitive to tuning parameters, we introduce additional tight hard constraints to guarantee that no mid-air collisions will occur.  The proposed method assumes that each agent is broadcasting its position and velocity on a common network. Additionally, to increase the avoidance robustness, we use the state estimator uncertainty to shape the collision term in the cost function and the optimization constraints.

\begin{figure}[t]
	\centering
	\includegraphics[width=0.95\columnwidth, trim={3cm 3cm 5cm 2cm},clip]{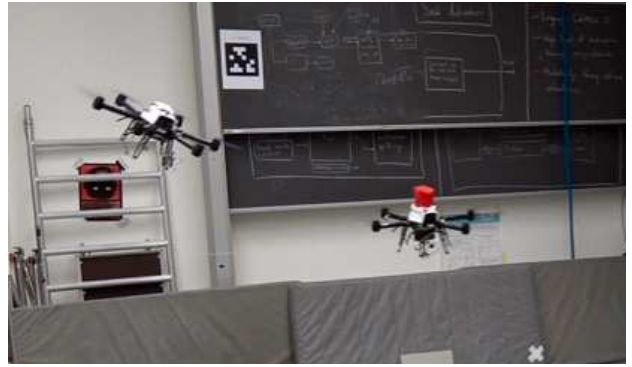}
	\caption{An instance of the experimental evaluation of multi-agent collision avoidance proposed strategy.}
	\label{pic:exp:masterslave}
\end{figure}

The contribution of this paper can be summarized as follows:
\begin{enumerate}[i]
	\item A unified framework for multi-agent control and collision avoidance.
	\item The incorporation of state estimator uncertainty and communication delay for robust collision avoidance.
\end{enumerate}

This paper is organized as follows: In Section~\ref{sec:related_work} we present an overview of existing methods for multi-agent collision avoidance. In Section~\ref{sec:model} we briefly present the \ac{MAV} model that will be considered in the controller formulation.  In Section~\ref{sec:controller_formulation} we present the controller and discuss the state estimator uncertainty propagation. Finally, in Section~\ref{sec:evaluation} we present simulation and experimental results of the proposed approach.



	\section{RELATED WORK}\label{sec:related_work}
	Many researchers have demonstrated successful trajectory generation and navigation on \acp{MAV} in controlled environment where obstacles are static, using external motion capture system \cite{michael2010grasp} or using on-board sensing \cite{burri2015real}.
Sampling based planning techniques can be used to generate global collision-free trajectories for single agent, taking into account the agent dynamics \cite{frazzoli2002real} and static and dynamics obstacles in the environment. 

One way to generate collision-free trajectories for a team of robots is to solve a mixed integer quadratic problem in a centralized fashion as shown in \cite{kushleyev2013towards}. A similar approach was presented in \cite{augugliaro2012generation} where sequential quadratic programming techniques are employed to generate collision-free trajectories for a team of \acp{MAV}. The aforementioned methods lack real-time performance and do not consider unforeseen changes in the environment.

Global collision-free trajectory generation methods limit the versatility of the team of robots. In real missions, where multiple agents are required to cooperate, the task assigned to each agent might change according to the current situation, and reactive local trajectory planning methods become crucial.  

One of the earliest works to achieve reactive collision avoidance for a team of flying robots in a \ac{NMPC} framework is the work presented in \cite{shim2003decentralized} where a decentralized \ac{NMPC} is employed to control multiple helicopters in a complex environment with an artificial potential field to achieve reactive collision avoidance. This approach does not provide any guarantees and has been evaluated only in simulation. 
In \cite{alonso2015collision} the authors present various algorithms based on the \ac{VO} concept to select collision-free trajectories from a set of candidate trajectories. The method has been experimentally evaluated on $4$ \acp{MAV} flying in close proximity and including human. However, the \ac{MAV} dynamics are not considered in this method, and decoupling trajectory generation from control has various limitations as shown in the experimental section of \cite{alonso2015collision}.  

Among the attempts to unify trajectory optimization and control is the work presented in \cite{neunert2016fast}. The robot control problem is formulated as a finite horizon optimal control problem and an unconstrained optimization is performed at every time step to generate time-varying feedback gains and feed-forward control inputs simultaneously. The approach has been successfully applied on \acp{MAV} and a ball balancing robot.

In this work, we unify the trajectory tracking and collision avoidance into a single optimization problem in a decentralized manner. In this way, trajectories generated from a global planner can be sent directly to the trajectory tracking controller without modifications, leaving the local avoidance task to the tracking controller.


	\section{MODEL}\label{sec:model}
	In this section we present the MAV model employed in the controller formulation. We first introduce the full vehicle model and explain the forces and moment acting on the system. Next, we will briefly discuss the closed-loop attitude model employed in the trajectory tracking controller.

\subsubsection{System model}
\begin{figure}
	\centering
	\resizebox{0.99\columnwidth}{!}{%
		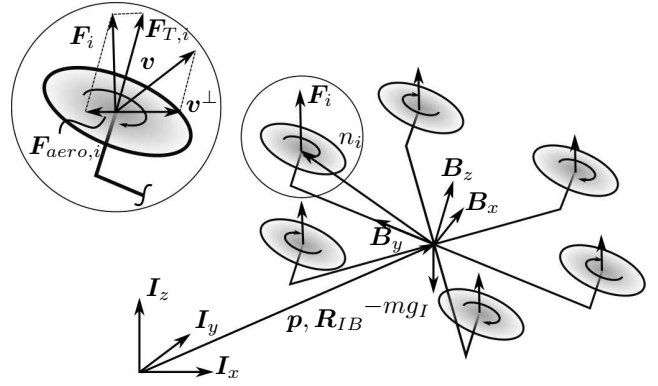
	}%
	\caption{A schematic of \ac{MAV} showing Forces and torques acting on the \ac{MAV} and aerodynamic forces acting on a single rotor. Inertial and \ac{CoG} frames are also shown.}
	\label{fig:mav_frames}
\end{figure}

We define the world fixed inertial frame $ I $ and the body fixed frame $ B $ attached to the \ac{MAV} in the \ac{CoG} as shown in Figure~\ref{fig:mav_frames}. The vehicle configuration is described by the position of the \ac{CoG} in the inertial frame $ \bm{p} \in \mathbb{R}^{3} $, the vehicle velocity in the inertial frame $ \bm{v} $, the vehicle orientation  $ \bm{R}_{I B}\in SO(3) $  which is parameterized by Euler angles and the body angular rate $ \bm{\omega} $.

The main forces acting on the vehicle are generated from the propellers. Each propeller generates thrust proportional to the square of the propeller rotation speed $ n_{i} $and angular moment due to the drag force. The generated thrust $ \bm{F}_{T,i} $ and moment $ \bm{M}_{i} $ from the $ i-th $ propeller is given by:

\begin{subequations}
	\begin{align}
	\bm{F}_{T,i} &= k_{n} n_{i}^{2} \bm{e}_z, 				\label{eq:prop_force}\\
	\bm{M}_{i} &= (-1)^{i-1} k_{m} \bm{F}_{T,i}, 		\label{eq:prop_moment}
	\end{align}
\end{subequations}
\noindent where $ k_n $ and $ k_m $ are positive constants and $\bm{e}_z$ is a unit vector in $z$ direction.
Moreover, we consider two important effects that appear in the case of dynamic maneuvers. These effects are the blade flapping and induced drag. The importance of these effects stems from the fact that they introduce additional forces in the $ x-y $ rotor plane, adding some damping to the \ac{MAV} velocity as shown in \cite{6289431}. It is possible to combine these effects as shown in \cite{6696696, 7487627} into one lumped drag coefficient $k_D$.

\noindent This leads to the aerodynamic force $\bm{F}_{aero,i}$:
\begin{equation}
\bm{F}_{aero,i} =  f_{T,i}\bm{K}_{drag}\bm{R}_{IB}^{T}\bm{v}
\end{equation}
where $  \bm{K}_{drag} = diag(k_D,k_D,0)$ and $f_{T,i}$ is the $z$ component of the $ i-th $ thrust force.

%

%

The motion of the vehicle can be described by the following equations:
\normalsize
\begin{subequations}
	\begin{eqnarray}
	\dot{\bm{p}} &=& \bm{v} ,\label{eq:dynamics_eq1} \\
	\dot{\bm{v}} &=& \frac{1}{m}\left( \bm{R}_{I B} \sum_{i=0}^{N_r}\bm{F}_{T,i} - \bm{R}_{IB}\sum_{i=0}^{N_r} \bm{F}_{aero,i}  + \bm{F}_{ext}\right) \nonumber \\ 
	& +& \left[ \begin{array}{ccc}
	0 \\ 
	0 \\ 
	-g
	\end{array}  \right],  \label{eq:dynamics_eq2} \\
	\dot{\bm{R}}_{IB} &=& \bm{R}_{IB} \lfloor \bm{\omega} \times \rfloor,  \label{eq:dynamics_eq3}\\
	\bm{J} \dot{\bm{\omega}}& = & -\bm{\omega} \times \bm{J} \bm{\omega} + \bm{\mathcal{A}} \left[ \begin{array}{c}
	n^2_{1} \\ 
	\vdots \\ 
	n^2_{N_{r}}
	\end{array}  \right], \label{eq:dynamics_eq4}
	\end{eqnarray}
\end{subequations}
\normalsize
\noindent where $ m $ is the mass of the vehicle and $ \bm{F}_{ext} $ is the external forces acting on the vehicle (i.e wind). $ \bm{J} $ is the inertia matrix, $ \mathbf{\mathcal{A}}  $ is the allocation matrix and $N_{r}$ is the number of propellers.

\subsubsection{Attitude model}
We follow a cascaded approach as described in \cite{blosch2010vision} and assume that the vehicle attitude is controlled by an attitude controller.
For completeness we quickly summarize their findings in the following paragraph.

To achieve accurate trajectory tracking, it is crucial for the high level controller to consider the inner loop system dynamics.
Therefore, it is necessary to consider a simple model of the attitude closed-loop response.
These dynamics can either be calculated by simplifying the closed loop dynamic equations (if the controller is known) or by a simple system identification procedure in case of an unknown attitude controller (on commercial platforms for instance).
In this work we used the system identification approach to identify a first order closed-loop attitude response. 

The inner-loop attitude dynamics are then expressed as follows:

\begin{subequations}
	\begin{align}
	&\dot{\phi} = \frac{1}{\tau}_{\phi} \left(k_{\phi} \phi_{cmd} - \phi  \right), \label{eq:roll_dynamics} \\ 
	&\dot{\theta} = \frac{1}{\tau}_{\theta} \left( k_{\theta}\theta_{cmd} - \theta \right),\label{eq:pitch_dynamics}\\ 
	&\dot{\psi} = \dot{\psi}_{cmd},  \label{eq:yaw_dynamics}
	\end{align}
	\label{eq:attitude_innerloop}
\end{subequations}
\noindent where $ k_{\phi}, k_{\theta} $ and $ \tau_{\phi}, \tau_{\theta} $ are the gains and time constant of roll and pitch angles respectively. $ \phi_{cmd} $ and $ \theta_{cmd} $ are the commanded roll and pitch angles and $ \dot{\psi}_{cmd} $ is commanded angular velocity of the vehicle heading.

The aforementioned model will be employed in the subsequent trajectory tracking controller to account for the attitude inner-loop dynamics. Note that the vehicle heading angular rate $ \dot{\psi} $  is assumed to track the command instantaneously. This assumption is reasonable as the \ac{MAV} heading angle has no effect on the \ac{MAV} position.


	\section{CONTROLLER FORMULATION}\label{sec:controller_formulation}
	In this section we present the unified trajectory tracking and multi-agent collision avoidance \ac{NMPC} controller. First, we will present the \ac{OCP}. Afterwards we will discuss the cost function choice and the state estimator uncertainty propagation to achieve robust collision avoidance. Next, we will present the optimization constraints and finally we will discuss the approach adopted to solve the \ac{OCP} in real-time on-board of the \ac{MAV}.
\subsection{Optimal Control Problem}
To formulate the \ac{OCP}, we first define the system state vector $ \bm{x} $ and control input $ \bm{u} $ as follows:
\begin{eqnarray}\label{eq:state_input_definition}
\bm{x} = \left[ \begin{array}{ccccc}
\bm{p}^T & \bm{v}^T & \phi & \theta & \psi
\end{array} \right]^T  \\
\bm{u} = \left[ \begin{array}{ccc}
\phi_{cmd} & \theta_{cmd} & T_{cmd}
\end{array} \right] ^{T}
\end{eqnarray}

Every time step, we solve the following \ac{OCP} online:

\small
\begin{equation} \label{eq:mav_nonlinear_mpc_opt}
\begin{aligned}
\min_{\bm{U},\bm{X}} &\
\int_{t=0}^{T} \left\{ J_{\bm{x}}\left( \bm{x}(t), \bm{x}_{ref}(t) \right) + J_{\bm{u}} \left( \bm{u}(t), \bm{u}_{ref}(t) \right) + J_{c} \left( \bm{x}(t)\right)  \right\}dt \\ &+ J_{T}\left( \bm{x}(T)\right)  \\
&\begin{aligned}
\text{subject to} &
& & \dot{\bm{x}} = \bm{f}(\bm{x}, \bm{u});\\
& & & \bm{u}(t) \in \mathbb{U} \\
& & & \bm{G}(\bm{x}(t)) \leq 0 \\
& & & \bm{x}(0) = \bm{x}\left( {t_0}\right).
\end{aligned}
\end{aligned}
\end{equation}
\normalsize
\noindent where $ \bm{f} $ is composed of Equations~\eqref{eq:dynamics_eq1},~\eqref{eq:dynamics_eq2} and~\eqref{eq:attitude_innerloop}. $ J_{\bm{x}}, J_{\bm{u}}, J_{c} $ are  the cost function for reference trajectory $ \bm{x}_{ref} $ tracking, control input penalty and  collision cost function and  $ J_{T} $ is the terminal cost function. $\bm{G}$ is a function that represents the state constraint,  and $ \mathbb{U} $ is the set of admissible control inputs. In the rest of this section, we will discuss the details of the aforementioned \ac{OCP} and discuss a method to efficiently solve it in real-time. 
\subsection{Cost Function}
In this subsection we discuss the components of the cost function presented in \eqref{eq:mav_nonlinear_mpc_opt}. The first term $ J_{\bm{x}}\left( \bm{x}(t), \bm{x}_{ref}(t) \right)  $ penalizes the deviation of the predicted state $ \bm{x} $ from the desired state vector $ \bm{x}_{ref} $ in a quadratic sense as shown below:
\begin{equation}
J_{\bm{x}}\left( \bm{x}(t), \bm{x}_{ref}(t) \right)  = \left\|\bm{x}(t) - \bm{x}_{ref}(t)\right\|^{2}_{\bm{Q}_{x}} 
\end{equation}
\noindent where $ \bm{Q}_{x} \succeq 0 $ is a tuning parameter. The state reference $ \bm{x}_{ref} $ is obtained from the desired trajectory. 
The second term in the cost function is related to the penalty on the control input as shown below:
\begin{equation}
J_{\bm{u}} \left( \bm{u}(t), \bm{u}_{ref}(t) \right)  = \left\|\bm{u}(t) - \bm{u}_{ref}(t)\right\|^{2}_{\bm{R}_{u}}
\end{equation}
\noindent where $  \bm{R}_{u} \succeq 0 $ is a tuning parameter. The control input reference $ \bm{u}_{ref} $ is chosen to achieve better tracking performance based on desired trajectory acceleration as described in \cite{kamelmpc2016}.

\begin{figure}
	\centering
	\includegraphics[width=0.99\columnwidth]{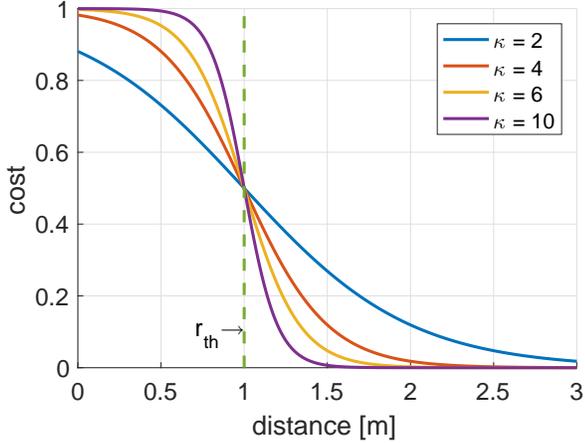}
	\caption{Logistic function based potential field for different smoothness parameter $ \kappa $.}
	\label{fig:LogisticCostFunction}
\end{figure}

The collision cost $  J_{c} \left( \bm{x}(t)\right)  $  to avoid collisions with other $ N_{agents} $ is given by:
\begin{equation}\label{eq:logistic_function}
\begin{aligned}
	 J_{c} \left( \bm{x}(t)\right)  &= \sum_{j=1}^{N_{agents}}  \frac{Q_{c,j}}{1 + \exp{\kappa_{j} \left( d_{j}(t) - r_{th, j}(t)\right) }} \\ &\text{for } j = 1, \dots, N_{agents}
\end{aligned}
\end{equation}
\noindent where $ d_{j}(t) $ is the Euclidean distance to the $ j-th $ agent given by $ d_{j}(t) = \left\| \bm{p}(t) - \bm{p}_{j}(t)\right\|_2  $,  $  Q_{c,j}  > 0 $ is a tuning parameter, $ \kappa_{j}  > 0$ is a parameter that defines the smoothness of the cost function and $ r_{th, j} (t)$ is a threshold distance between the agents  where the collision cost is $  Q_{c,j} / 2$. Equation~\eqref{eq:logistic_function} is based on the logistic function, and the main motivation behind this choice is to achieve a smooth and bounded collision cost function. Figure~\ref{fig:LogisticCostFunction} shows the cost function for different $ \kappa $ parameters.
%
%
\subsection{Constraints}
The first constraint in \eqref{eq:mav_nonlinear_mpc_opt} guarantees that the state evolution respects the \ac{MAV} dynamics. To compensate for external disturbances and modeling errors to achieve offset-free tracking, we employ a model-based filter to estimate external disturbances $ \bm{F}_{ext} $ as described in details in \cite{kamelmpc2016}.  The second constraints addresses limitations on the control input as follows: 
\small
\begin{equation}\label{eq:input_constraints}
\mathbb{U} = \left\lbrace \bm{u} \in \mathbb{R}^{3} | \left[ \begin{array}{c}
\phi_{min} \\ 
\theta_{min} \\ 
T_{cmd,min}
\end{array} \right] \leq \bm{u} \leq   \left[ \begin{array}{c}
\phi_{max} \\ 
\theta_{max} \\ 
T_{cmd,max}
\end{array} \right] \right\rbrace .
\end{equation}
\normalsize

The third constraint guarantees collision avoidance by setting tight hard constraints on the distance between two agents. The $ j- th $ row of the $ \bm{G} $ matrix represents the collision constraints with the $ j-th $ agent. This is given by:
\begin{equation}
\bm{G}_{j}(\bm{x}) = - \left\| \bm{p}(t) - \bm{p}_{j}(t) \right\|_2^2 + r^{2}_{min, j}(t) .
\end{equation}
\noindent where $  \bm{p}_{j}(t) $ is the position of the $ j-th $ agent at time $ t $. These are non-convex constraints and $ \bm{G} $ is continuous and smooth. $ r_{min, j} $ is chosen to always be strictly less than $ r_{th, j} $ to guarantee that the hard constraints are activated only if the potential field in Equation~\eqref{eq:logistic_function} is not able to maintain $ r_{min, j} $ distance to the $ j-th $ agent.

Finally, the last constraint in the optimization problem is to fix the initial state $ \bm{x}(0) $ to the current estimated state $ \bm{x}(t_0) $.

\subsection{Agents Motion Prediction}
Given that the approach presented in this work is based on Model Predictive Control, it is beneficial to employ a simple model for the other agents and use it to predict their future behavior. The model we employ in this work is based on a constant velocity model, but this can be replaced with more sophisticated model, and we will consider this for future work. Given the current position and velocity of the $ j-th $ agent $ \bm{p}_{j}(t_{0}), \bm{v}_{j}(t_{0}) $ we can predict the future positions of the $ j-th $ agent along the prediction horizon as follows:
\begin{equation}\label{eq:agent_motion_model}
\bm{p}_{j}(t) = \bm{p}_{j}(t_{0}) + \bm{v}_{j}(t_{0}) \left( t - t_{0}  + \delta \right) .
\end{equation} 
\noindent where $ \delta $ is the communication delay that we compensate for to achieve better prediction. $ \delta $ is calculated based on the difference between the timestamp on the message and the arrival time. This is possible thanks to a clock synchronization between the agents and a time server. The communication delay compensation can be omitted if there is no clock synchronization between agents.  Additionally, to reduce the noise sensitivity, we consider the velocity to be zero if it is below a certain threshold $ v_{th} $.

\subsection{Uncertainty Propagation}
\begin{figure}
	\centering
	\huge 
	\resizebox{0.99\columnwidth}{!}{%
		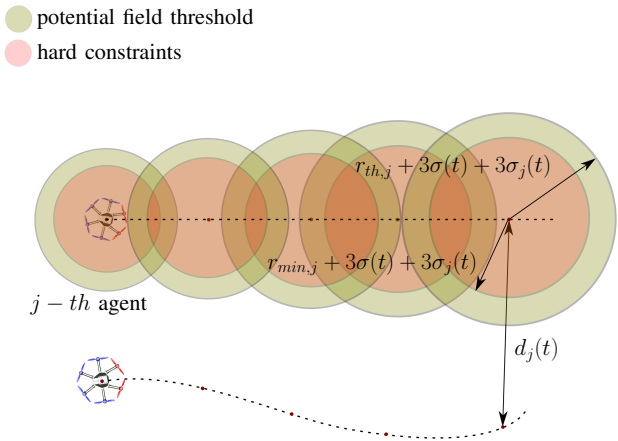
	}%
	\caption{The concept of robust collision avoidance, the minimum acceptable distance between two agents is increased along the prediction horizon based on the state uncertainty propagation. Tighter hard constraints will be activated only if the potential field fails to maintain the minimum acceptable distance $ r_{min, j} $. Note that the plotted trajectories are predicted trajectories obtained from the solution of the \ac{OCP}~\eqref{eq:mav_nonlinear_mpc_opt}.}
	\label{fig:robust_collision_avoidance_concept}
\end{figure}

To account for the uncertainty in the state estimator and the uncertainty of the other agents, to achieve higher level of robustness, we propagate the estimated state uncertainty to calculate the minimum allowed distance to the $ j-th $ agent $ r_{min, j}(t)$  and the threshold distance $ r_{th, j} (t)$. In other words, if the state is highly uncertain, we should be more conservative on allowing agents to get closer to each other by increasing $ r_{min, j} $ and $ r_{th, j} $ at time $ t $ along the prediction horizon. The uncertainty of the $ j-th $ agent's position is propagated using the model described in Equation~\eqref{eq:agent_motion_model}, while the self-uncertainty can be propagated with higher accuracy employing the system model described in Equations~\eqref{eq:dynamics_eq1},~\eqref{eq:dynamics_eq2} and~\eqref{eq:attitude_innerloop}. In many previous works, the uncertainty propagation is typically performed using the unscented transformation when the system is nonlinear. In our case, given that we need real-time performance, we choose to perform uncertainty propagation based on an \ac{EKF}. Given the current predicted state $ \bm{x}(t_{0}) $ with covariance $ \bm{\Sigma}(t_0) $, we propagate the uncertainty by solving the following differential equation:
\begin{equation}\label{eq:uncertainty_propagation}
\dot{\bm{\Sigma}}(t) = \bm{F}(t) \bm{\Sigma}(t) \bm{F}(t)^{T} 
\end{equation}
\noindent with boundary condition $ \bm{\Sigma}(0) = \bm{\Sigma}(t_0) $. $ \bm{F}(t) $ is the state transition Jacobian matrix. Using Equation~\eqref{eq:uncertainty_propagation} we compute the $ j-th $ agent's uncertainty and the self-uncertainty at time $ t $, namely $ \bm{\Sigma}_{j}(t)  $ and $ \bm{\Sigma}(t) $. These values are employed to calculate $ r_{min, j} (t) $ and $ r_{th, j}(t) $ according to the following Equations:
\begin{equation}
\begin{aligned}
r_{min, j} (t)&= r_{min}+ 3\sigma(t) + 3\sigma_{j}(t),\\
r_{th, j}(t) &= r_{th} + 3\sigma(t) + 3\sigma_{j}(t).
\end{aligned}
\end{equation}
\noindent where $ \sigma $ is the square root of the maximum eigenvalue of the self-uncertainty $ \bm{\Sigma} $ and $ \sigma_{j} $ is the square root of maximum eigenvalue of the $ j-th $ agent's uncertainty $ \bm{\Sigma}_{j} $. $ r_{min} $ and $ r_{th} $ are constant parameters.
We use the maximum eigenvalue to reduce the problem of computing the distance between two ellipsoids, which is more complex and time consuming, to the computation of the distance between two spheres. Approximating the uncertainty ellipsoid by the enclosing sphere makes the bounds more conservative, especially if the uncertainty is disproportionately large only in a particular direction.
Figure~\ref{fig:robust_collision_avoidance_concept} illustrates the concept for 2 agents.

\subsection{Implementation}
\emph{A Multiple shooting} technique \cite{Kirches2011} is employed to solve \eqref{eq:mav_nonlinear_mpc_opt}. The system dynamics and constraints are discretized over a coarse discrete time grid $ t_{0} , \dots, t_{N}$ within the time interval $ \left[ t_{k}, t_{k+1} \right]  $. For each interval, a \ac{BVP} is solved, where  additional continuity constrains are imposed. An implicit \emph{RK} integrator of order $ 4 $ is employed to forward simulate the system dynamics along the interval.
At this point, the \ac{OCP} can be expressed as a \ac{NLP} that can be solved using \ac{SQP} technique where an active set or interior point method can be used to solve the \ac{QP}. For the controller to behave as a local planner and to guarantee problem feasibility, a long prediction horizon $ T $ is necessary. To achieve this without significantly increasing the computation effort, the time step of the grid over which the system dynamics is discretized is chosen to be larger than the controller rate. Smooth predictions are obtained thanks to the implicit Runge-Kutta of order 4 integrator employed and because the system dynamics is not represented by stiff differential equations.

\subsection{Priority}
In many situations, a particular agent may have high priority to follow the reference trajectory. For instance, if a \ac{MAV} is delivering an object or taking footage for mapping task. The proposed method can handle priorities simply by changing the connectivity graph between agents. The highest priority \ac{MAV}  will not perform any avoidance maneuver and therefore doesn't need to have access to other agents state. A lower priority agent will perform avoidance maneuver only to avoid higher priority agents.


	\section{EVALUATION}\label{sec:evaluation}
	In this section we evaluate the proposed approach in simulation and experimentally. The proposed controller has been implemented in C++ and integrated into \ac{ROS}. The ACADO toolkit \cite{Houska2011a} is employed to generate a fast C solver for the \ac{OCP}.
First we show two  simulation studies in a high fidelity \acp{MAV} simulator, RotorS \cite{Furrer2016} where $ 6 $ \acp{MAV} are commanded to swap positions simultaneously. Then we present two experimental evaluations of the method with two \acp{MAV}.

\subsection{Simulation Results}
\begin{figure*}[b]
	\centering
	\begin{subfigure}[b]{0.99\textwidth}
		\centering
		\includegraphics[width=0.92\textwidth, trim={4cm 1cm 2cm 0cm},clip]{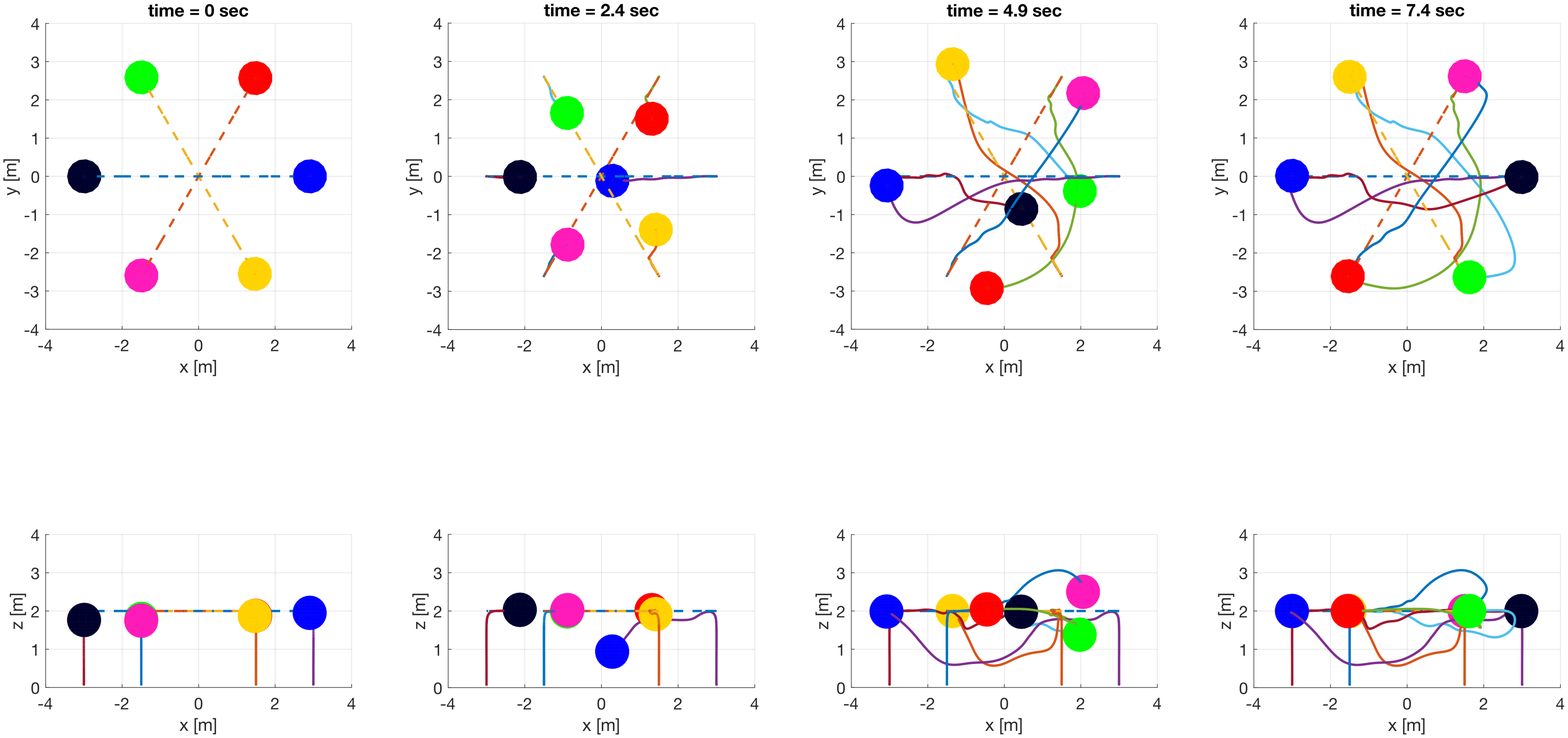}
		\caption{Trajectories of 6 \acp{MAV} during a position swapping simulation. The \acp{MAV} are represented by spheres, while dotted lines represent the reference trajectory provided to the controller. Solid lines represent the actual trajectory executed  by each agent.}
		\label{fig:sim_6mavs_reciprocal}
	\end{subfigure}\\
	\begin{subfigure}[b]{0.99\textwidth}
		\centering
		\includegraphics[width=0.92\textwidth, trim={4cm 1cm 2cm 0cm},clip]{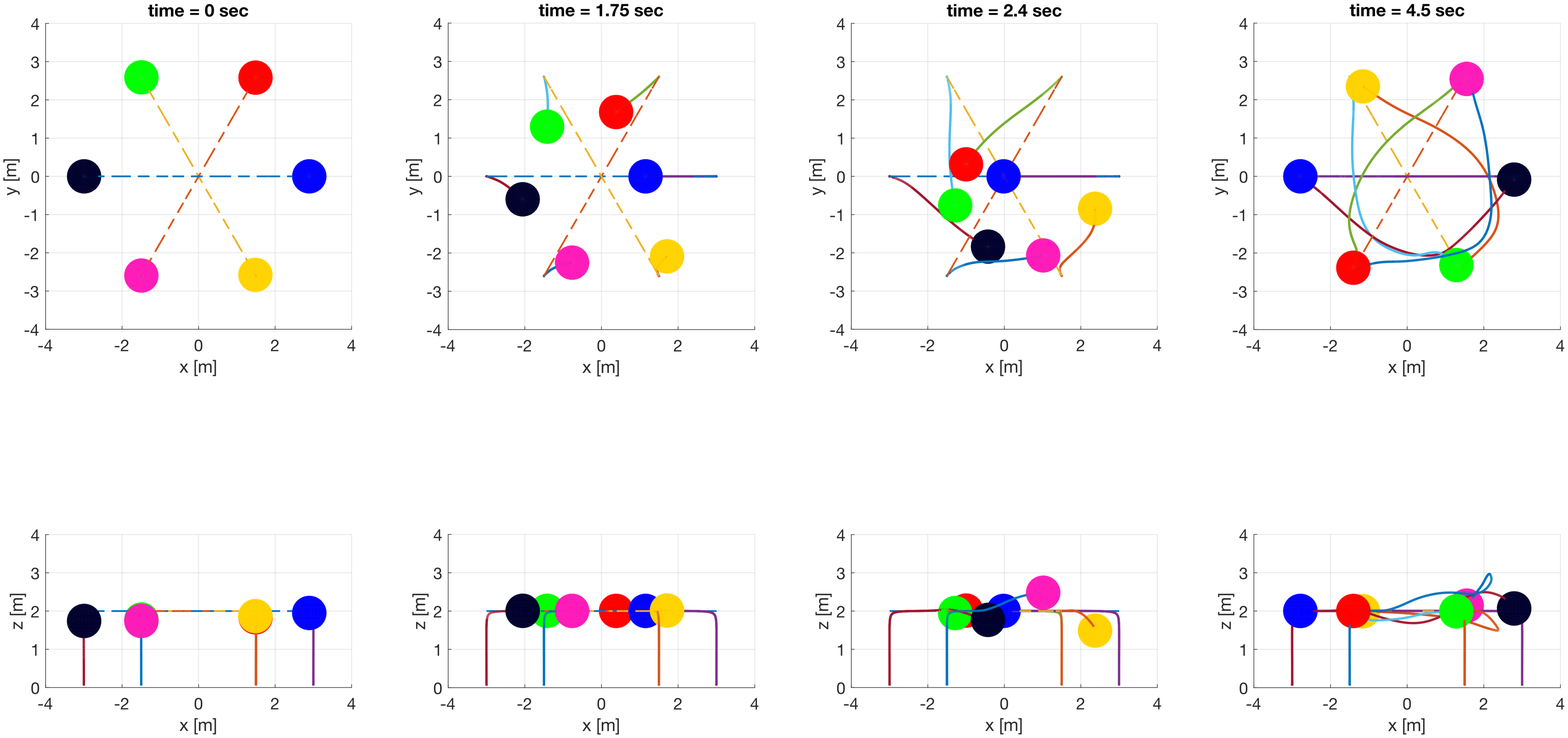}
		\caption{Trajectories of 6 \acp{MAV}  during a position swapping simulation with priority given to the first \ac{MAV} (blue sphere). Dotted lines represent the reference trajectory provided to the controller. Solid lines represent the actual trajectory executed by each agent.}
		\label{fig:sim_6mavs_priority}
	\end{subfigure}
	\caption{Simulation results of 6 \acp{MAV} exchanging positions}
	\label{fig:sim_results}
\end{figure*}

In this simulation study we evaluate our method on $ 6 $ \acp{MAV} commanded to exchange their initial positions. The reference trajectory for each agent is not collision-free. Figure~\ref{fig:sim_6mavs_reciprocal} shows the evolution of the \ac{MAV} trajectories when all agents are sharing position and velocity information in a fully connected graph. In this case, avoidance maneuvers are reciprocal. The average computation time during this simulation is \unit[1.0]{ms} while worst case computation time is around \unit[4.0]{ms} on an Intel i7 \unit[2.8]{GHz} CPU.


In another simulation study depicted in Figure~\ref{fig:sim_6mavs_priority}, we assigned hierarchical avoidance scheme, where the first \ac{MAV} (blue sphere) has top priority to follow the reference trajectory, while the second \ac{MAV} (red sphere) is avoiding only the first \ac{MAV}. The third \ac{MAV} is avoiding the first two, etc... .

Clearly, assigning priority scheme makes the problem simpler to solve with better position swapping trajectories as shown in Figure~\ref{fig:sim_results}.

%

%


%

\subsection{Experimental Results}
We experimentally evaluated the proposed approach on two AscTec NEO hexacopters. In a first experiment, one \ac{MAV} is commanded to hover in position, while a second \ac{MAV} with high priority is commanded manually to fly in vicinity to the first \ac{MAV}. In a second experiment, crossing reference trajectories are planned for both \acp{MAV} and executed. The purpose of these experiments is to show how the proposed method exploits the system dynamics and abruptly responds to changes in the other agent behavior in a robust manner.

\subsubsection{Experimental Setup}
The AscTec NEO hexacopter is equipped with an Intel i7 \unit[2.8]{GHz}  \unit[8]{GHz} RAM computer running \ac{ROS} and an onboard flight controller and as well as tuned attitude controller. The on-board computer communicates with the flight controller over a serial port and exchanges information at \unit[100]{Hz}. An external motion capture system (Vicon) is used to measure each \ac{MAV} pose. This measurement is fused with the onboard \ac{IMU} to achieve accurate state estimation using the multi-sensor fusion framework \cite{lynen2013robust}. The estimated position and velocity of each agent is shared over the network. The controller is running onboard of each \ac{MAV} at \unit[100]{Hz} while the prediction horizon of the controller is chosen to be \unit[2]{seconds}.

\subsubsection{Experiments}
\begin{figure*}[htp]
	\centering
	\begin{subfigure}[t]{0.95\columnwidth}
	\centering
	\includegraphics[width=0.99\columnwidth, height=1.4\columnwidth]{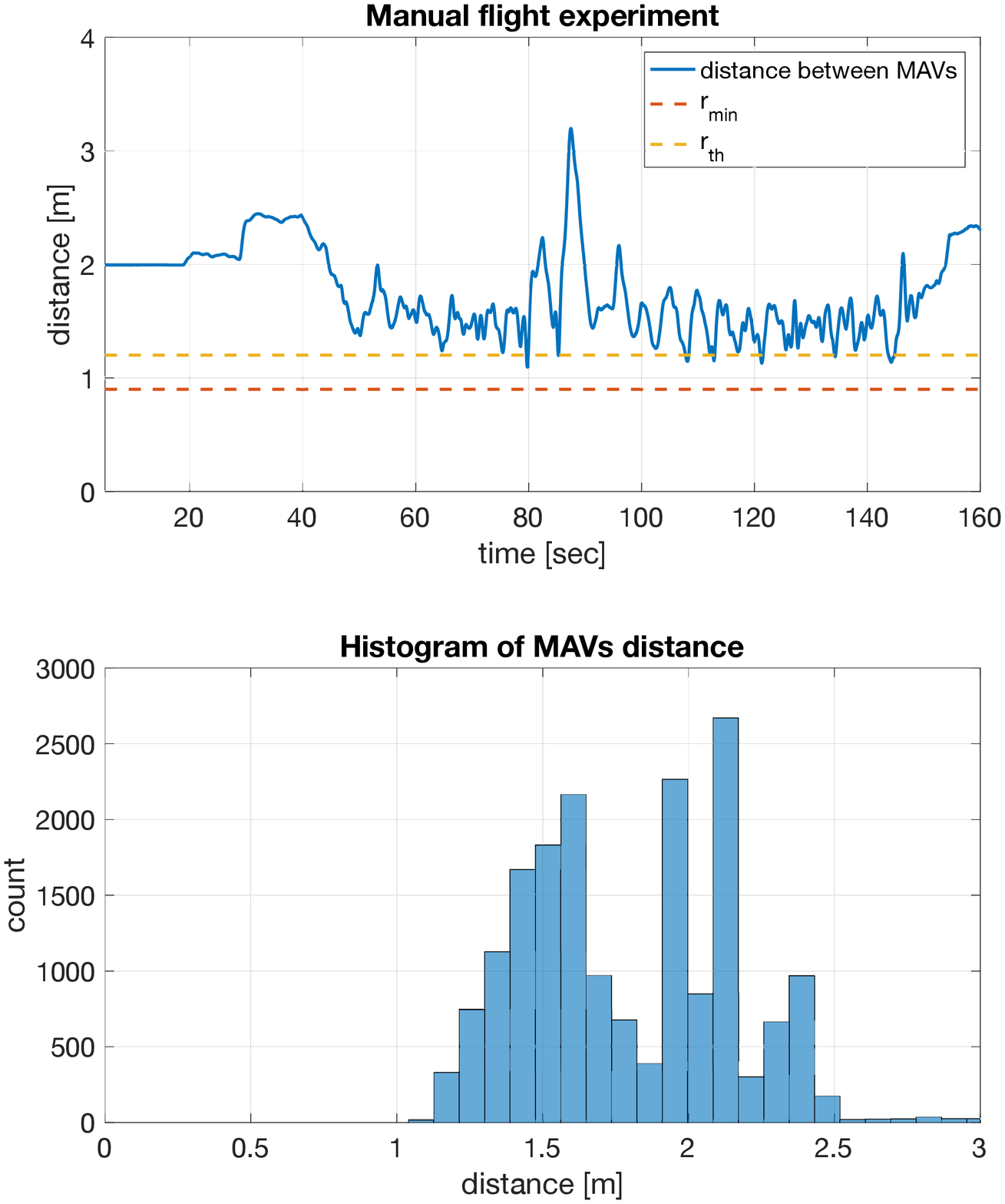}
	\caption{The upper plot shows the distance between two \acp{MAV} over time during the manual flight experiment. One \ac{MAV} is commanded to hover in position while the other one is manually commanded to approach it with a priority assigned to the manually commanded \ac{MAV}. $ r_{th} $ is set to  \unit[1.2]{m} while the hard constraint on the distance is set to $ r_{min} = $ \unit[0.9]{m}. The lower plot shows the histogram of the distance. }
	\label{fig:exp_manual_flight}
\end{subfigure} \hfill
\begin{subfigure}[t]{0.95\columnwidth}
	\centering
	\includegraphics[width=0.99\columnwidth, height=1.4\columnwidth]{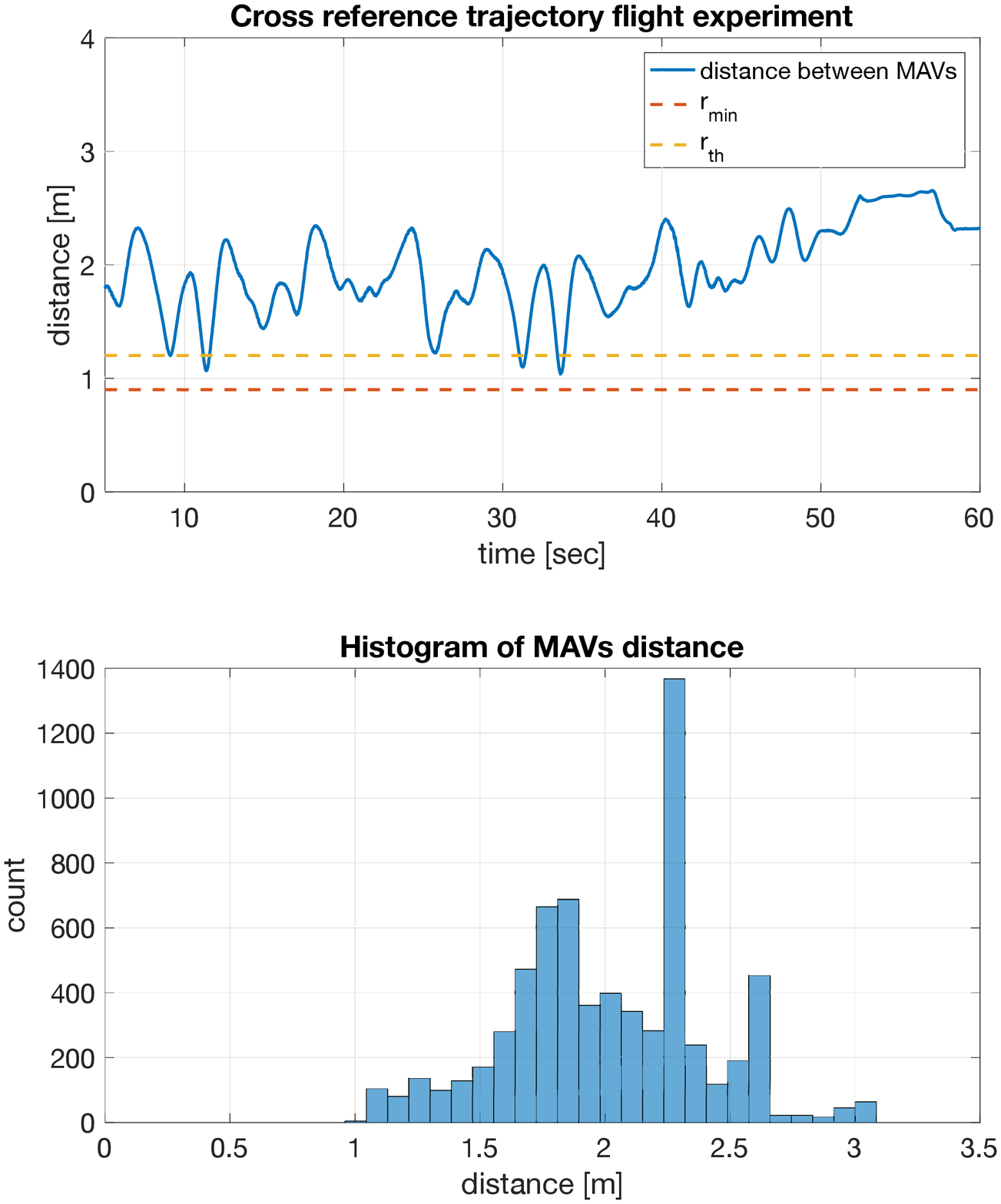}
	\caption{The upper plot shows the distance between two \acp{MAV} over time during the crossing trajectories experiment. $ r_{th} $ is set to  \unit[1.2]{m} while the hard constraint on the distance is set to $ r_{min} = $ \unit[0.9]{m}. The lower plot shows the histogram of the distance. }
	\label{fig:exp_cross_traj}
\end{subfigure}
\label{fig:exp_results}
\caption{Experimental results with two \acp{MAV}}
\end{figure*}

Figure~\ref{fig:exp_manual_flight} shows the distance between the two agents over time. The distance is almost always maintained above $ r_{th} $ except for moments of aggressive maneuvers, however the hard constraint was never activated as the distance was always above $ r_{min} $ all the time.

\begin{figure*}
	\centering
	\includegraphics[width=0.99\textwidth, trim={4cm 0cm 4cm 0cm},clip]{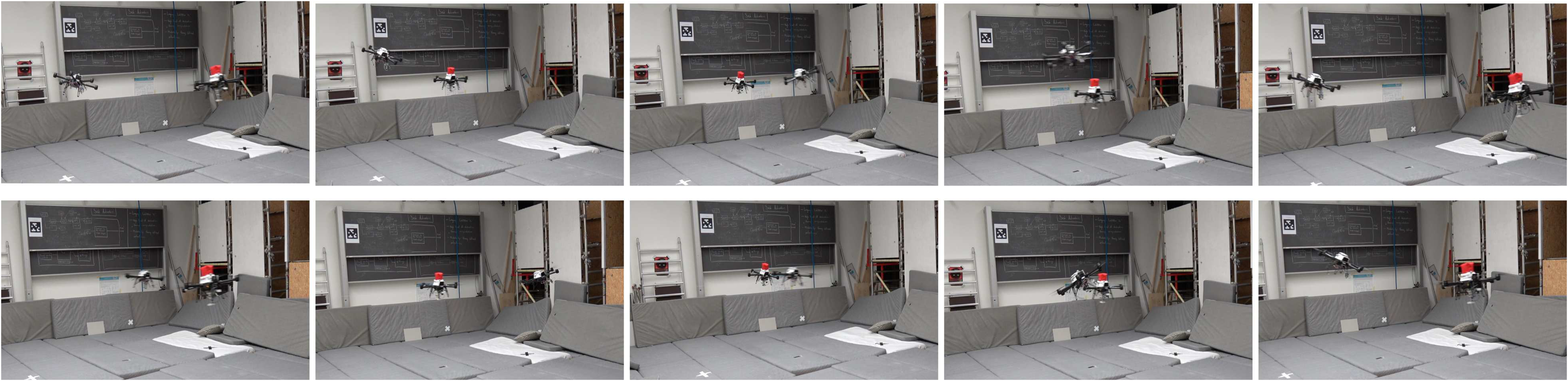}
	\caption{A sequence of images during the cross trajectories experiments. The two \acp{MAV} are commanded to follow a non collision-free trajectories with priority assigned to the \ac{MAV} with the red hat. Images are taken  \unit[1]{second} apart starting from top left.}
	\label{fig:cross_traj_seq}
\end{figure*}

In the second experiment, intersecting reference trajectories with maximum velocity of \unit[2]{m/s} are planned and sent as reference trajectories to the respective \ac{MAV}. Figure~\ref{fig:cross_traj_seq} shows a sequence of images taken \unit[1]{second} apart during the experiment \footnote{video available on \url{http://goo.gl/RWRhmJ}}. Figure~\ref{fig:exp_cross_traj} shows the distance between the two \acp{MAV} over time. The hard constraint was never active since the distance was always above $ r_{min} $.

	\section{CONCLUSIONS}\label{sec:conclusion}

	In this paper we presented a multi-\acp{MAV} collision avoidance strategy based on Nonlinear Model Predictive Control. The approach accounts for state estimator uncertainty by propagating the uncertainty along the prediction horizon to increase the minimum acceptable distance between agents, providing robust collision avoidance. Tight hard constraints on the distance between agents guarantee no collisions if the prediction horizon is sufficiently long. Moreover, by changing the connectivity graph, it is possible to assign priority to certain agents to follow their reference trajectories. The approach has been evaluated in simulation with 6 agents and in real experiments with 2 agents. Our experiments showed that this collision avoidance approach results into agile and dynamic avoidance maneuvers while maintaining system stability at reasonable computational cost.


	\section*{ACKNOWLEDGMENT}
	This work was supported by the  European Union's Horizon 2020 Research and Innovation Programme under the Grant Agreement No.644128, AEROWORKS.

		\begin{acronym}
			\acro{MAV}{Micro Aerial Vehicle}
			\acro{NMPC}{Nonlinear Model Predictive Control}
			\acro{MPC}{Model Predictive Controller}
			\acro{CoG}{Center of Gravity}
			\acro{OCP}{Optimal Control Problem}
			\acro{SQP}{Sequential Quadratic Programming}
			\acro{NLP}{Nonlinear Program}
			\acro{BVP}{Boundary Value Problem }
			\acro{VO}{Velocity Obstacles}
			\acro{EKF}{Extended Kalman Filter}
			\acro{QP}{Quadratic Program}
			\acro{ROS}{Robot Operating System}
			\acro{IMU}{Inertial Measurement Unit}

		\end{acronym}

	\bibliographystyle{IEEEtran}

	\bibliography{./bibliography/bibliography}

	\twocolumn
	\addtolength{\textheight}{-12cm}   




\end{document}